\documentclass[conference]{IEEEtran}
\IEEEoverridecommandlockouts
\usepackage{cite}
\usepackage{amsmath,amssymb,amsfonts}
\usepackage{algorithmic}
\usepackage{graphicx}
\usepackage{textcomp}
\usepackage{xcolor}

\usepackage{algorithm}  
\usepackage{caption}

\def\BibTeX{{\rm B\kern-.05em{\sc i\kern-.025em b}\kern-.08em
    T\kern-.1667em\lower.7ex\hbox{E}\kern-.125emX}}
\begin{document}

\title{DFR: DEPTH FROM ROTATION BY UNCALIBRATED IMAGE RECTIFICATION
WITH LATITUDINAL MOTION ASSUMPTION
\thanks{This work is supported by the National Key R\&D Program of China (No. 2022ZD0119003), Nature Science Foundation of China (No. 62102145), and Jiangxi Provincial 03 Special Foundation and 5G Program (Grant No. 20224ABC03A05).}
}

\author{Yongcong Zhang$^{1,\ast}$\thanks{*Both authors contributed equally to this research.} \qquad Yifei Xue$^{2,\ast}$ \qquad Ming Liao$^{2}$ \qquad Huiqing Zhang$^{1}$ \qquad Yizhen Lao$^{1,\dag}$\thanks{\dag Corresponding author:yizhenlao@hnu.edu.cn.}
\\
\\
$^{1}$ Hunan University\\$^{2}$ 
Jiangxi Provincial Natural Resources Cause Development Center}

\maketitle

\begin{abstract}
Despite the increasing prevalence of rotating-style capture (e.g., surveillance cameras), conventional stereo rectification techniques frequently fail due to the rotation-dominant motion and small baseline between views. In this paper, we tackle the challenge of performing stereo rectification for uncalibrated rotating cameras. To that end, we propose Depth-from-Rotation (DfR), a novel image rectification solution that analytically rectifies two images with two-point correspondences and serves for further depth estimation.
Specifically, we model the motion of a rotating camera as the camera rotates on a sphere with fixed latitude. The camera's optical axis lies perpendicular to the sphere's surface. We call this latitudinal motion assumption. Then we derive a 2-point analytical solver from directly computing the rectified transformations on the two images. We also present a self-adaptive strategy to reduce the geometric distortion after rectification. Extensive synthetic and real data experiments demonstrate that the proposed method outperforms existing works in effectiveness and efficiency by a significant margin.
\end{abstract}


\section{Introduction}
\label{sec:intro}

\label{sec:intro}

Image rectification is vital for efficient stereo matching by forcing the point correspondence restricted in the same scan line (row). This process significantly reduces the computational cost for further depth estimation and thus is widely used in 3D vision applications, such as robotics, autonomous driving, and augmented reality.\\

\noindent \textbf{Motivations.} Classical image rectification techniques apply homographies on a pair of images (i.e., the master and the slave image) whose epipolar geometry is pre-computed. Thus the epipolar lines in the original images map to horizontally aligned lines in the transformed images~\cite{hartley2003multiple}. 

\begin{figure}[t]
    \centering
    \includegraphics[width=1\linewidth]{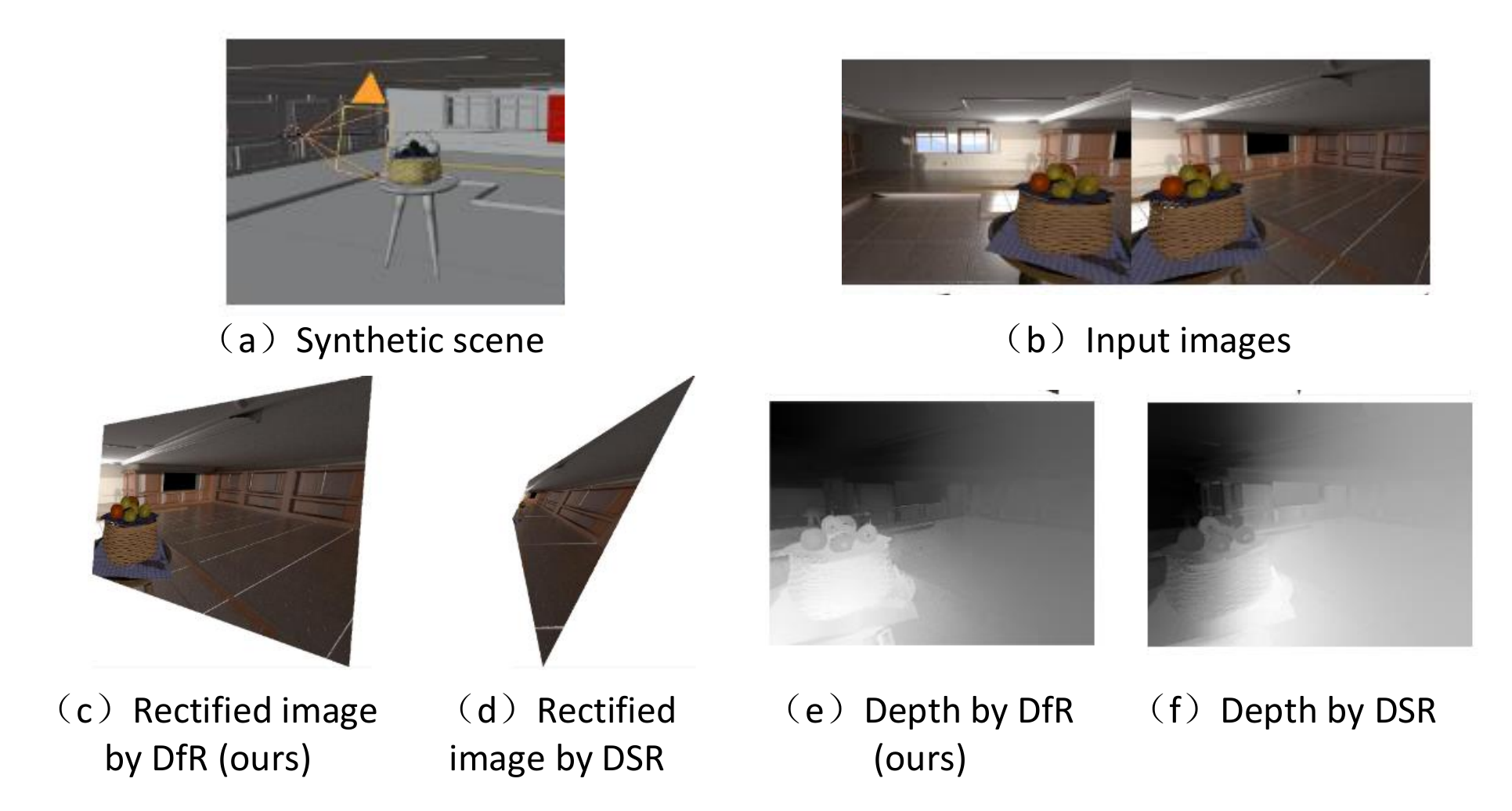}
    \caption{A rotating camera (a) captures two images (b). Stereo rectified results and estimated depth map by proposed method \textit{DfR} (c)(e) and state-of-the-art work \textit{DSR}~\cite{xiao2018dsr} (d)(f).}
    \label{fig:intro}
\end{figure}

\begin{figure}[t]
    \centering
    \includegraphics[width=.8\linewidth]{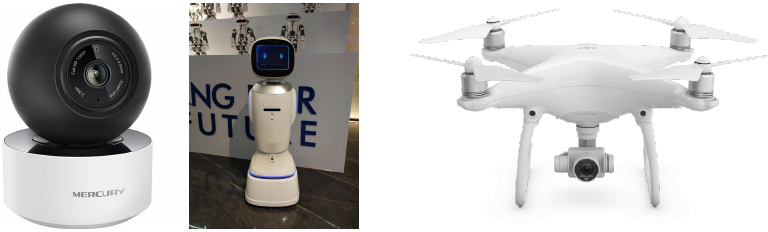}
    \caption{Typical rotating camera systems.}
    \label{fig:example}
\end{figure}

However, we found the ubiquitous 2D rotating cameras, such as surveillance cameras and tripod head cameras on UAVs (Fig.~\ref{fig:example}), fail in applying conventional image rectification for the three shortcomings shared in the existing works (Fig.~\ref{fig:intro}):  

\textit{\textbf{1) Cumbersome calibration.}} The works of \cite{fusiello2000compact,ventura2016structure} require off-line intrinsic calibration and have to fix the camera setting, such as focal length. However, such cumbersome and stringent requirements are challenging to maintain in real-world applications. 

\textit{\textbf{2) Poor epipolar geometry estimation with short baseline.}} Another type of rectification methods~\cite{loop1999computing,sweeney2019structure} demand estimated epipolar geometry as input. Note that accurate epipolar computation is highly dependent on establishing a sufficient baseline between images so that the rectification can be reliably estimated. However, the rotating  cameras produce extremely short baselines that violate such an assumption.  

\textit{\textbf{3) Over-distorted rectification.}} Existing solutions could lead to significant geometric distortion among rectified images once the slave and master images have significant relative rotation, which is not preferred for high-quality depth-based applications~\cite{xiao2018dsr}. \\

\noindent \textbf{Contributions.}  This paper tackles the challenge of stereo rectification for uncalibrated rotating cameras with extremely short baselines. To this end, we propose a novel image rectification solution that analytically computes the rectified transformations to the two images with only two-point correspondences and serves for the further depth estimation, short for Depth-from-Rotation (\textbf{DfR}). Specifically, we model the motion of a rotating camera as the camera rotates on a sphere with fixed latitude. The camera's optical axis lies perpendicular to the sphere's surface. We call this \textbf{"latitudinal motion assumption"}. Then we derive a 2-point analytical solver that computes the rectified transformations for the two images directly. We also present a self-adaptive strategy to reduce the geometric distortion after rectification. 

Previous research~\cite{ventura2016structure} presents structure from motion (SfM) solution with spherical motion, which is similar to our latitudinal motion. However, we want to highlight that~\cite{ventura2016structure} requires accurate pre-calibration of intrinsic while the proposed \textbf{DfR} is a calibration-free method. Besides, the radius of sphere in~\cite{ventura2016structure} is assumed as brachium ($\approx0.8m$) while with only a centimeter length for rotating camera case. The most related work ~\cite{xiao2018dsr} rectifies uncalibrated cameras with slight movements. Nevertheless, such a method assumes a small but pure translation by neglecting the rotation. In contrast, we consider both rotation and translation produced by latitudinal motion. 

Differing from previous methods, our contributions are: 

\noindent $\bullet$ We describe the rotating camera as a latitudinal motion and find that rectified transformations for such a case can be computed directly and efficiently with only two-point correspondences.

\noindent $\bullet$ Extensive experiments demonstrate that the proposed rectification method \textbf{DfR} outperforms state-of-the-art techniques in align accuracy and distortion suppression. The code has been uploaded to GitHub\footnote{https://github.com/zhangtaxue/DFR}.

\section{Related Work}

Finding the epipolar geometry is the critical step in the previous approach to launching the rectification. A detailed review of the related techniques is in~\cite{zhang1998determining}. Typically, \cite{loop1999computing} proposes to compute the fundamental matrix first and then extract the homographies followed by a decomposing to reduce the geometric distortion. 
Gluckman and Nayar~\cite{gluckman2001rectifying} present a rectification approach that
minimizes the re-sampling effect. \cite{isgro1999projective} introduces an implicit image rectification approach by computing homographies directly from point correspondences. 
\cite{fusiello2011quasi} alternatively utilizes a Quasi-Euclidean-based rectification algorithm. 
More recently, \cite{xiao2018dsr} proposes a direct uncalibrated image rectification solution for the monocular camera with a small translation. 

As opposed to existing methods that require calibrated stereo rig, estimated epipolar geometry, or pure translation assumption, we propose a practical solution called DfR to rectify two uncalibrated images captured by a rotating camera with an extremely short baseline. 

\section{Methodology}

\begin{figure}[t]
    \centering
    \includegraphics[width=1\linewidth]{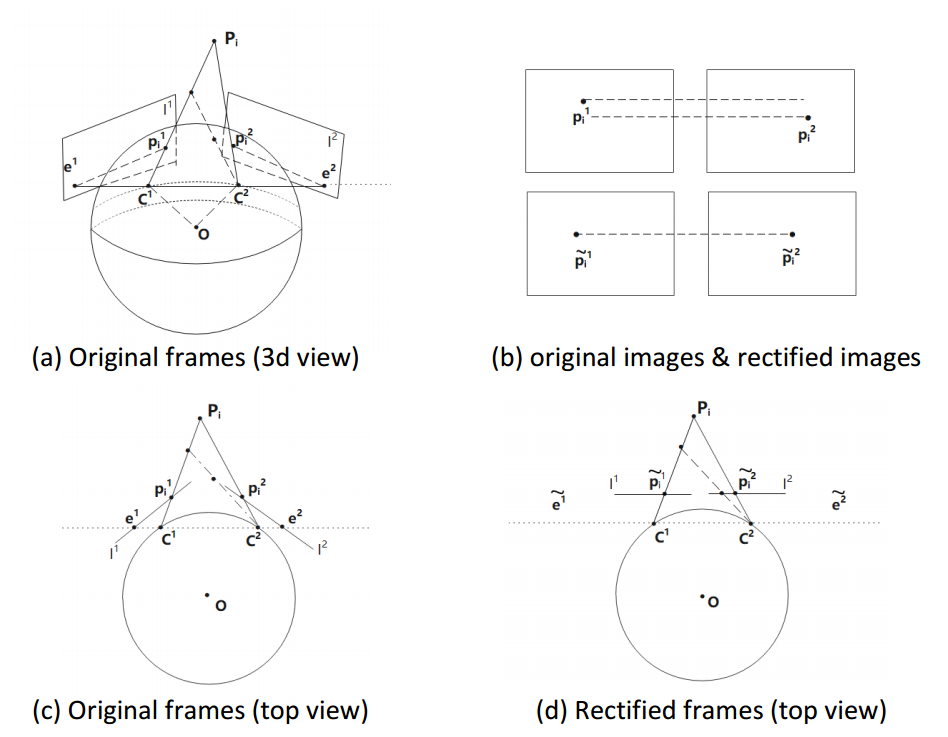}
    \caption{3D view of general latitudinal motion (a). The objective of our rectification is to shift the original frames (c) to the rectified frames (d), this process also map the original images to the rectified ones (b). }
    \label{fig:illustrate1}
\end{figure}

\subsection{The Geometry of Latitudinal Camera Motion}

We highlight that the rotating camera, such as a surveillance camera, follows a "latitudinal camera motion" trajectory. With this assumption, the camera rotates at a constant latitude $\overset{\frown}{{C^1} {C^2}}$ with a fixed distance from an origin, and the optical axis is aligned with the ray between the origin and camera center $\overrightarrow{\left | O{C^1} \right |}$ (Fig.~\ref{fig:illustrate1}(a)(c)). 

Assuming a 3D point $\mathbf{P}_i=[X_i,Y_i,Z_i]^\top$ is captured by $C^1$ and $C^2$ as $\mathbf{p}_i^1=[x_i^1,y_i^1]$ and $\mathbf{p}_i^2=[x_i^2,y_i^2]$:

\begin{equation}
    \alpha_i [\mathbf{p}_i^j,1]^\top = \mathbf{K}^j[\mathbf{R}^j|\mathbf{t}^j][\mathbf{P}_i,1]^\top
\label{equation:projection_model}
\end{equation}

\noindent where $\mathbf{R}^j\in SO(3)$ and $\mathbf{t}^j\in \mathbb{R}^3$ are rotation matrix and translation vector of $j^{\text{th}}$ camera. $\alpha_i$ is a scalar w.r.t depth.  
$\mathbf{K}$ is a $3 \times 3$ matrix known as the calibration matrix containing the intrinsic parameters of a camera. Since the majority of the existing works assume the principal points are usually also close to the image center and $\mathbf{K}$ keeps constant between two views~\cite{hartley2003multiple}, we adhere to this assumption too in this paper. Thus $\mathbf{K} = \mathbf{K}^1=\mathbf{K}^2$ are defined as: 

\begin{equation}
    \mathbf{K} = \text{diag}(f_x,f_y,1)
    \label{equation:K_matrix_define}
\end{equation}

\noindent where $f_x$ and $f_y$ are focal lengths along x and y axes.

\subsection{Finding the Rectified Matrices}
\label{section:H_computation}

\begin{figure}[t]
    \centering
    \includegraphics[width=1\linewidth]{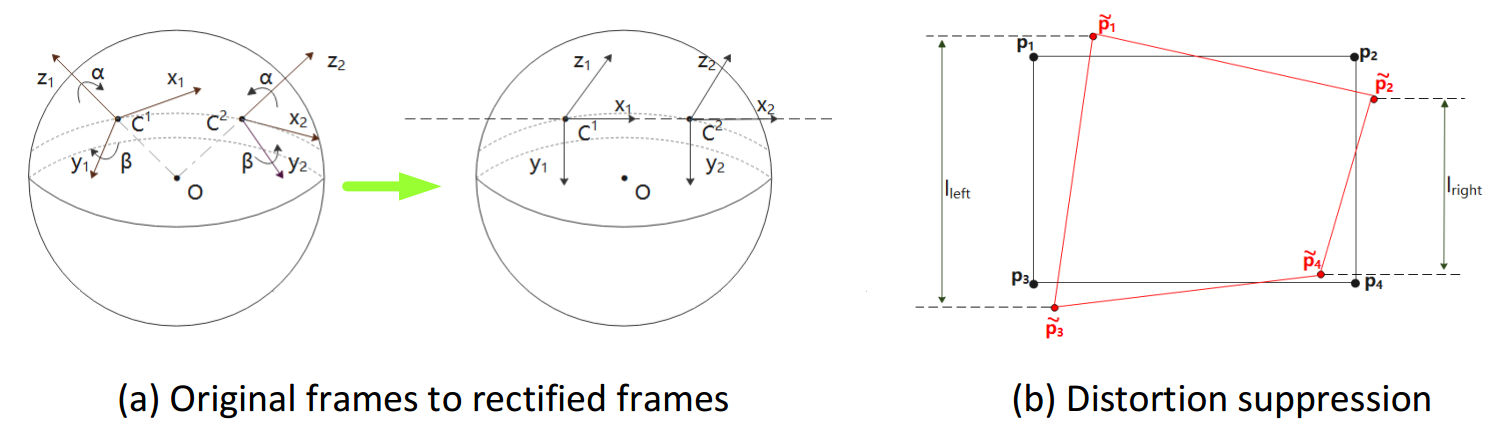}
    \caption{(a) The transformation of camera coordinate system between original and rectified frames. (b) Geometric distortion suppression. }
    \label{fig:illustrate2}
\end{figure}

The primary aim of the rectification that we seek is to transform the stereo under latitudinal camera motion into a laterally displaced stereo setting (fig.~\ref{fig:illustrate1}(d)). We present a novel rectification solution which leads to such transformation. 

Specifically, we derive the rotation-based rectification (Sect.~\ref{section:rotation_rectification}) and introduce the analytical solver for rectified matrices (Sect.~\ref{section:2pt_solver}). We also present a self-adaptive strategy to suppress the geometric distortion (Sect.~\ref{section:distortion_suppress}) and integrate these steps into a complete rectification pipeline  (Sect.~\ref{section:pipeline}).   
\\
\subsubsection{Derivation of Rectified Matrices to $C^1$ and $C^2$}\label{section:rotation_rectification}\quad\\

\noindent We show that stereo $C^1\leftrightarrow C^2$ under latitudinal camera motion can be rectified to laterally displaced stereo $\tilde{C}^1 \leftrightarrow \tilde{C}^2$ by applying two homographies $\mathbf{H}_1$ and $\mathbf{H}_2$ to  $C^1$ and $C^2$ respectively. \\ 

\noindent \textbf{Proposition 1.} When two views  $C^1$ and $C^2$ have a constant but unknown focal length and under latitudinal camera motion, after applying the $\mathbf{H}_1$ and $\mathbf{H}_2$ defined as 

\begin{small}
\begin{equation}\label{equ:H1}
\mathbf{H}^1=\left[
\begin{array}{rcl}
\cos(\beta)\cos(\alpha) & -\frac{f_{x}\sin(\alpha)\sin(b)}{f_{y}} & -f_{x}\cos(\alpha)\sin(\beta)\\
\frac{f_{y}cos(\beta)sin(\alpha)}{f_{x}}  & cos(\alpha) & -f_{y}\sin(\alpha)\sin(\beta)\\
\frac{\sin(\beta)}{f_{x}} & 0  & \cos(\beta)
\end{array} \right] 
\end{equation}
\end{small}
\begin{small}
\begin{equation}\label{equ:H2}
\mathbf{H}^2=\left[
\begin{array}{rcl}
\cos(\beta)\cos(\alpha) & \frac{f_{x}\sin(\alpha)\sin(b)}{f_{y}} & f_{x}\cos(\alpha)\sin(\beta)\\
-\frac{f_{y}\cos(\beta)\sin(\alpha)}{f_{x}}  & \cos(\alpha) & -f_{y}sin(\alpha)sin(\beta)\\
-\frac{\sin(\beta)}{f_{x}} & 0  & \cos(\beta)
\end{array} \right]
\end{equation}
\end{small}

\noindent to $C^1$ and $C^2$ respectively, they will be transform to  $\tilde{C}^1$ and $\tilde{C}^2$ that under perfect laterally displace. Where $\alpha$ and $\beta$ are two angles that control the rotations of $C^1$ and $C^2$. \\

\noindent \textit{Proof.} As shown in Fig.~\ref{fig:illustrate2}(a), we assume the frames of $C^1$ and $C^2$ are  w.r.t world coordinate system $O$. Thus, with latitudinal camera motion assumption, the poses of $C^1$ and $C^2$ are: 

\begin{align}
[\mathbf{R}^1|\mathbf{t}^1] = [\mathit{R}(\alpha,-\beta,0),\mathbf{z}] \label{equation:C1_pose}\\
[\mathbf{R}^2|\mathbf{t}^2] = [\mathit{R}(-\alpha,\beta,0),\mathbf{z}]  \label{equation:C2_pose}
\end{align}

\noindent where $\mathbf{z}=[0,0,-1]^\top$ and $\mathit{R}(\alpha,\beta,\gamma) = R_z(\alpha )R_y(\beta )R_x(\gamma)$ is the matrix multiplication of atomic matrices whose yaw, pitch, and roll angles are $\alpha$,$\beta$, $\gamma$. Thus, $\mathbf{R}^1$ and $\mathbf{R}^2$ are: 

\begin{small}
\begin{equation}\label{equ:R1}
\mathbf{R}^1=\left[
\begin{array}{rcl}
cos(\beta)cos(\alpha) & sin(\alpha) & cos(\alpha)sin(\beta)\\
-cos(\beta)sin(\alpha) & cos(\alpha) &  -sin(\beta)sin(\alpha)\\
-sin(\beta) & 0 & cos(\alpha)
\end{array} \right] 
\end{equation}
\end{small}

\begin{small}
\begin{equation}\label{equ:R2}
\mathbf{R}^2=\left[
\begin{array}{rcl}
cos(\beta)cos(\alpha) & -sin(\alpha) & -cos(\alpha)sin(\beta)\\
cos(\beta)sin(\alpha) & cos(\alpha) &  -sin(\beta)sin(\alpha)\\
sin(\beta) & 0 & cos(\alpha)
\end{array} \right] 
\end{equation}
\end{small}

Obviously, we can rotate $C^1$ and $C^2$ to coincide with the world frame by rotation matrices ${\mathbf{R}^1}^{-1}$ and ${\mathbf{R}^2}^{-1}$. Note that with such a setting, the two novel views are laterally displaced stereo. Thus, the homographies which can perform the rectification are:

\begin{equation}
\mathbf{H}^1=\mathbf{K}{\mathbf{R}^1}^{-1}\mathbf{K}^{-1} \qquad \mathbf{H}^2=\mathbf{K}{\mathbf{R}^2}^{-1}\mathbf{K}^{-1} 
\label{equation:H_define}
\end{equation}

Finally, by subsisting Eq.~(\ref{equ:R1}), Eq.~(\ref{equ:R2}) and Eq.~(\ref{equation:H_define}), we can obtain Eq.~(\ref{equ:H1}) and Eq.~(\ref{equ:H2}). \hfill $\Box$
\\
\subsubsection{Computation of $\mathbf{H}^1$ and $\mathbf{H}^2$}
\label{section:2pt_solver}\quad\\

\noindent \textbf{$\bullet$ Decomposition of $\mathbf{H}^1$ and $\mathbf{H}^2$.}
Follows the homograph decomposition introduced in~\cite{loop1999computing}, we express the rectified homographies in 
Proposition 1 as the matrix multiplication of two $3\times3$ matrices:   

\begin{equation}
    \mathbf{H}^1 = \mathbf{H}_s^1 \mathbf{H}_y^1 \qquad \mathbf{H}^2 = \mathbf{H}_s^2 \mathbf{H}_y^2
\end{equation}

\noindent where, 
\begin{equation}
    \mathbf{H}_s^1 = \begin{bmatrix}
S_a & S_b & 0\\ 
0 & 1 & 0\\ 
0 & 0 & 1
\end{bmatrix} \qquad
\mathbf{H}_s^2 = \begin{bmatrix}
S_a & -S_b & 0\\ 
0 & 1 & 0\\ 
0 & 0 & 1
\end{bmatrix}
\end{equation}

\begin{equation}
\label{equation:H_y_defind}
    \mathbf{H}_y^1 = \begin{bmatrix}
1 & 0 & 0\\
h_{21} & h_{22} & h_{23}\\
h_{31} & 0 & h_{33}
\end{bmatrix} 
\mathbf{H}_y^2 = \begin{bmatrix}
1 & 0 & 0\\
-h_{21} & h_{22} & h_{23}\\
-h_{31} & 0 & h_{33}
\end{bmatrix}
\end{equation}

We use $\mathbf{H}^1_y$ and $\mathbf{H}^2_y$  to align the  correspondences along the y-axis while $\mathbf{H}^1_s$ and $\mathbf{H}^2_s$ serve to reduce the geometric distortion of the transformed images. \\

\noindent \textbf{$\bullet$ Computation of $\mathbf{H}^1_y$ and $\mathbf{H}^2_y$.} Notice that the elements of $\mathbf{H}^1_y$ and $\mathbf{H}^2_y$ share the same values  $h_{21}$, $h_{22}$, $h_{23}$, $h_{31}$, and $h_{33}$ but with different signs.
Thus, we can compute $\mathbf{H}^1_y$ and $\mathbf{H}^2_y$ by recovering the values of these elements. \\

    \noindent \textbf{Proposition 2.} Given two-point correspondences $\mathbf{p}_1^1\leftrightarrow \mathbf{p}_1^2$ and $\mathbf{p}_2^1\leftrightarrow \mathbf{p}_2^2$,  we first arbitrarily set the values of $h_{22}$ and $h_{23}$ and then obtain

\begin{equation}
h_{31} = \frac{t_{1}}{h_{22}}, \quad
h_{33} = \frac{1}{h_{22}}, \quad
h_{21}=t_{1}h_{23}+t_{2}h_{22}
\end{equation}

where \noindent $t_1$ and $t_2$ can be extracted by $\mathbf{t}=\mathbf{A}^{-1}\mathbf{b}$ with following definition: 

\begin{equation}
\underbrace{\begin{bmatrix}
-(x^{2}_{1}y^{1}_{1}+x^{1}_{1}y^{2}_{1}) &  (x^{1}_{1}+x^{2}_{1})\\
-(x^{2}_{2}y^{1}_{2}+x^{1}_{2}y^{2}_{2}) &  (x^{1}_{2}+x^{2}_{2})
\end{bmatrix}}_{\mathbf{A}} \underbrace{\begin{bmatrix}
t_1\\ 
t_2
\end{bmatrix}}_{\mathbf{T}} = \underbrace{\begin{bmatrix}
-y^{1}_{1}+y^{2}_{1}\\
-y^{1}_{2}+y^{2}_{2} 
\end{bmatrix}}_{\mathbf{b}}
\label{equation:2pt_solver}
\end{equation}

\noindent where $\mathbf{A}$ and $\mathbf{b}$ are $2\times2$ matrix and $2\times1$ vector consisted by the coordinates of the two correspondences. \\

\noindent \textit{Proof.} By given a point correspondence $\mathbf{p}_i^1\leftrightarrow \mathbf{p}_i^2$, we  minimize
the vertical alignment error after applying $\mathbf{H}^1_y$ and $\mathbf{H}^2_y$ via 

\begin{equation}\label{equ:min}
[\mathbf{H}_{y}^{1}, \mathbf{H}_{y}^{2}] = \mathop{\arg\min}\sum_{i}(\frac{\mathbf{h}^{1}_{2}\mathbf{p}^{1}_{i}}{\mathbf{h}^{1}_{3}\mathbf{p}^{1}_{i}}-\frac{\mathbf{h}^{2}_{2}\mathbf{p}^{2}_{i}}{\mathbf{h}^{2}_{3}\mathbf{p}^{2}_{i}})^{2} 
\end{equation}

\noindent where $\mathbf{h}_x^y$ is the $x^{\text{th}}$ row of $\mathbf{H}^{y}$. Let

\begin{equation}
\frac{h^{1}_{2}p^{1}_{i}}{h^{1}_{3}p^{1}_{i}}-\frac{h^{2}_{2}p^{2}_{i}}{h^{2}_{3}p^{2}_{i}} = 0,
\label{equation:cost_func}
\end{equation}

\noindent we can substitute elements of $\mathbf{H}$ in Eq.~(\ref{equation:H_y_defind}) and coordinates of $\mathbf{p}_i$ into Eq.~(\ref{equation:cost_func})

\begin{small}
\begin{equation}\
    h_{22}h_{31}(-x^{2}_{i}y^{1}_{i}-x^{1}_{i}y^{2}_{i}) +(-h_{23}h_{31}+h_{21}h_{33})(x^{1}_{i}+x^{2}_{i})=(y^{2}_{i}-y^{1}_{i})
\end{equation}
\end{small}

\noindent where we force $h_{22}h_{33}=1$ since homography matrix is up to scale. Then with two-point correspondences $i=1,2$, we can get Eq.~(\ref{equation:2pt_solver}). \hfill $\Box$\\

\noindent \textbf{$\bullet$ Computation of $\mathbf{H}^1_s$ and $\mathbf{H}^2_s$.} We employ the method introduced in~\cite{loop1999computing} to compute $S_a$ and $S_b$. Since this section does not contain our contributions, we provide only the essential information for understanding the remainder of this paper. More details of the algorithm are available in ~\cite{loop1999computing,xiao2018dsr}. \\

\begin{figure}[t]
    \centering
    \includegraphics[width=.8\linewidth]{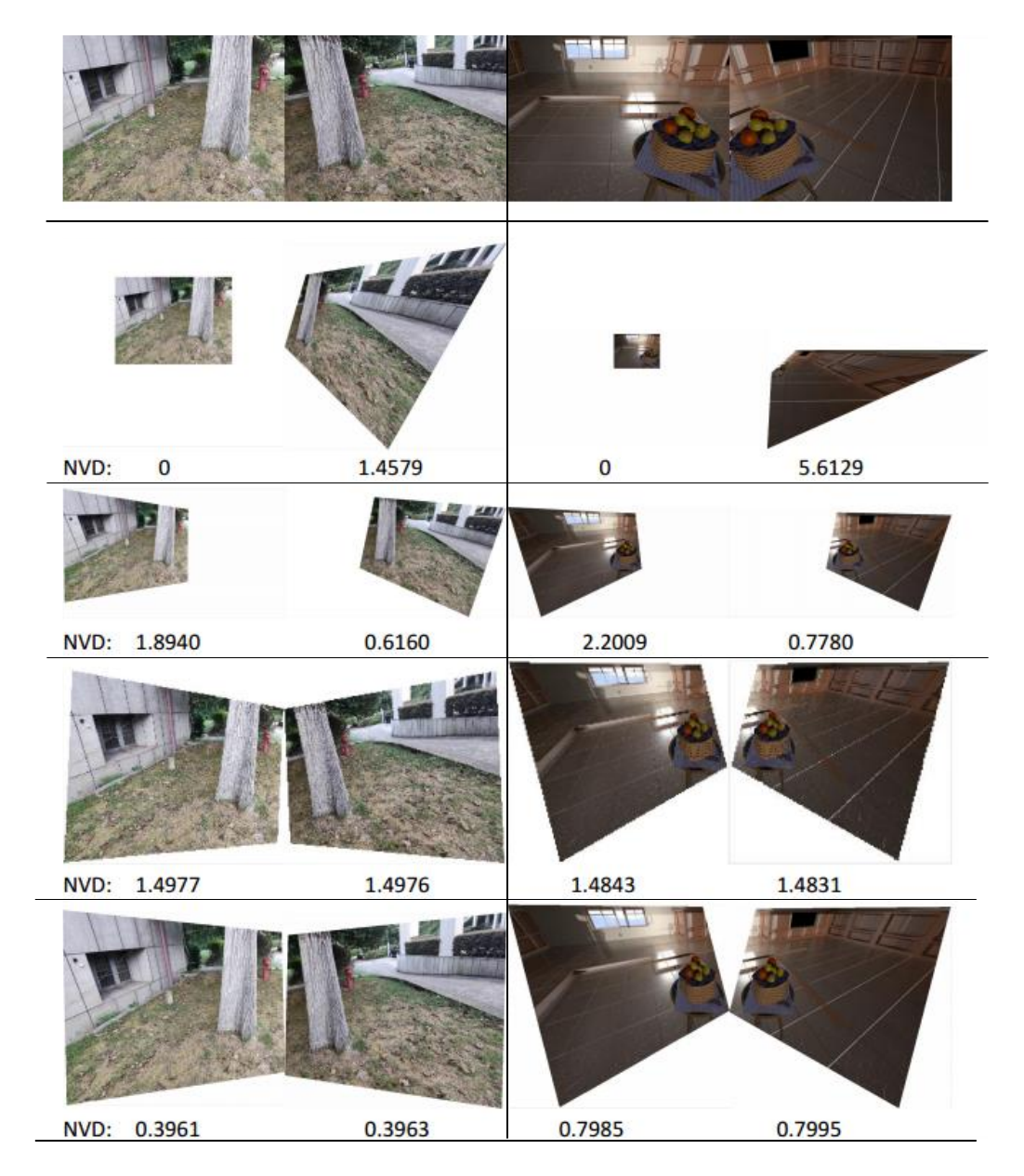}
    \caption{Ablation study on the geometric distortion suppression. The input images (1st row) are rectified by \textit{DSR}~\cite{xiao2018dsr} (2rd row), \textit{Loop}~\cite{loop1999computing} (3nd row),  proposed \textit{DfR} with (5th row) and without self-adaptive distortion suppression module (4th row). The NVD scores are given below for each result. }
    \label{fig:ablation}
\end{figure}

\subsubsection{Geometric Distortion Suppression}
\label{section:distortion_suppress}\quad\\

\noindent Recall that proposition 1 can hold with arbitrary value settings of $h_{22}$ and $h_{23}$. However, we point out that their values decide the geometric distortion level. Thus, a good value choice of $h_{22}$ and $h_{23}$ suppresses the distortion after rectification. 

Experimentally, we found out that the value of $h_{23}$ has a tiny inference about the shape of the output frame. Thus, it is critical to decide the value of $h_{22}$. As shown in Fig.~\ref{fig:illustrate2}(b), we assume the vertexes of original frame are $\mathbf{p_{1}}=[-\frac{W}{2},-\frac{H}{2},1]^{\top}$, $\mathbf{p_{2}}=[\frac{W}{2},-\frac{H}{2},1]^{\top}$, $\mathbf{p_{3}}=[-\frac{W}{2},\frac{H}{2},1]^{\top}$, and $\mathbf{p_{4}}=[\frac{W}{2},-\frac{H}{2},1]^{\top}$ ($H$ and $W$ are the height and width of image). After rectification by applying $\mathbf{H}^1$ and $\mathbf{H}^2$, the four vertexes becomes $\tilde{p_{1}}$, $\tilde{p_{2}}$, $\tilde{p_{3}}$, and $\tilde{p_{4}}$. Based on the description of $\mathbf{H}^1$ and $\mathbf{H}^2$in  Eq.~(\ref{equation:H_define}), we can measure the heights of the left edge and right edge as

\begin{equation}\label{equ:lenth}
\renewcommand*{\arraystretch}{1.5}
\left\{
\begin{array}{ccc}
l_{left} =  \tilde{p_{3}^{y}} - \tilde{p_{1}^{y}} = \frac{2Hh_{22}^{2}}{2-t_{1}W}-H \\
l_{right} =  \tilde{p_{4}^{y}} - \tilde{p_{2}^{y}} = \frac{2Hh_{22}^{2}}{2+t_{1}W}-H\\
\end{array} \right. 
\end{equation}

To reduce the geometric distortion, we force $l_{left}+l_{right} = 2H$. Thus, we have the instruction to set $h_{22}$:

\begin{equation}\label{equ:h22}
\begin{array}{c}
h_{22} = \sqrt{\frac{4-W^{2}t_{1}^{2}}{2}}
\end{array}
\end{equation}

\subsection{Pipeline}
\label{section:pipeline}

Note that we can only use two-point correspondences to perform rectification. However, we propose a RANSAC-like framework to robustly estimate the rectified matrices $\mathbf{H}^1$ and $\mathbf{H}^2$ by using the vertical alignment error $\textbf{VAE}$ between y coordinates of a rectified correspondence $\mathbf{\tilde{p_{l}}}=[\tilde{x_{l}}\ \tilde{y_{l}}]^{\top}$ and $\mathbf{\tilde{p_{r}}}=[\tilde{x_{r}}\ \tilde{y_{r}}]^{\top}$ 

\begin{equation}
\text{VAE}=\frac{\sum_{i=1}^n(|\tilde{y_{li}}-\tilde{y_{ri}}|)}{n}
\label{euqation:VAE}
\end{equation}

We describe the complete rectification pipeline in Alg.~\ref{alg:DfR}. 

\begin{algorithm}
	\renewcommand{\algorithmicrequire}{\textbf{Input:}}
	\renewcommand{\algorithmicensure}{\textbf{Output:}}
	\caption{DfR(Depth from Rotation)}
	\begin{algorithmic}[1]
 \label{alg:DfR}
		\REQUIRE Uncalibrated stereo images
		\STATE Matching correspondences:$\mathbf{p}_i^1\leftrightarrow \mathbf{p}_i^2$, $i=1..N$
		\STATE Initialize the vertical direction error: $\text{VER}_{min}$=inf
		\FOR{$t = 1 \ to \ T$}
		\STATE Randomly select point matches and compute temporary $\mathbf{\hat{H}}_y^1$ and $\mathbf{\hat{H}}_y^2$ using proposition 2 and Eq.~(\ref{equ:h22}). 
		\STATE Calculate VAE by applying $\mathbf{\hat{H}}_y^1$ and $\mathbf{\hat{H}}_y^2$ and take $\text{VAE}_{min} = min(\text{VER}_{min}, \text{VAE})$.
        \IF{$\text{VAE}<$  $\text{VAE}_{min}$} 
            \STATE {$\text{VAE}_{min}= \text{VER}$, $\mathbf{H_{y}^{1}}=\mathbf{\hat{H}}_y^1$, $\mathbf{H_{y}^{2}}=\mathbf{\hat{H}}_y^2$} 
        \ENDIF
		\ENDFOR
		\STATE Calculate $\mathbf{H_{s}^{1}}$ and $\mathbf{H_{s}^{2}}$ based on Sec.~\ref{section:H_computation}.
             \ENSURE $\mathbf{H}^1= \mathbf{H_{s}^{1}H_{y}^{1}}$ and $\mathbf{H}^2= \mathbf{H_{s}^{2}H_{y}^{2}}$
	\end{algorithmic}  
\end{algorithm}

\begin{figure*}[t]
    \centering
    \includegraphics[width=.95\linewidth]{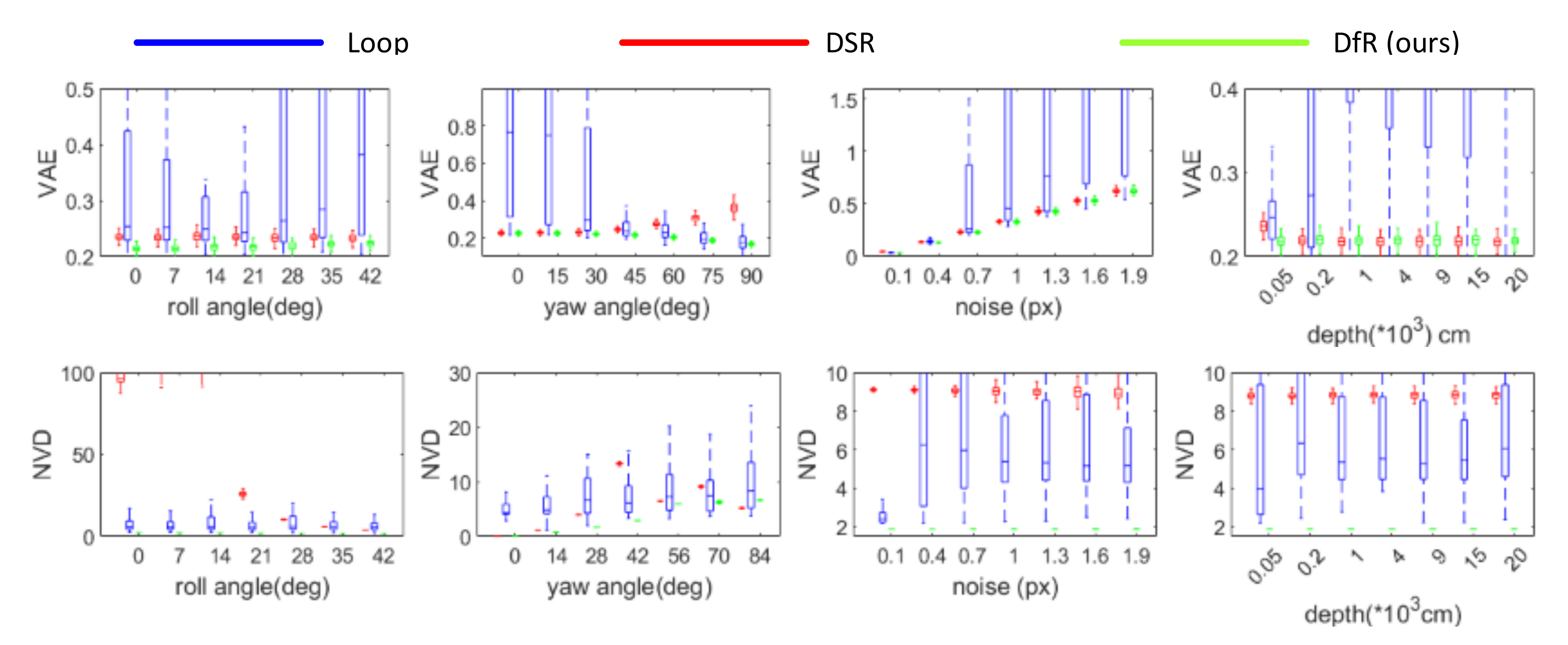}
    \caption{VAE and NVD scores of the rectification results produced by \textit{Loop}, \textit{DSR} and \textit{DfR} under different settings.}
    \label{fig:box}
\end{figure*}

\begin{figure}[t]
    \centering
    \includegraphics[width=1\linewidth]{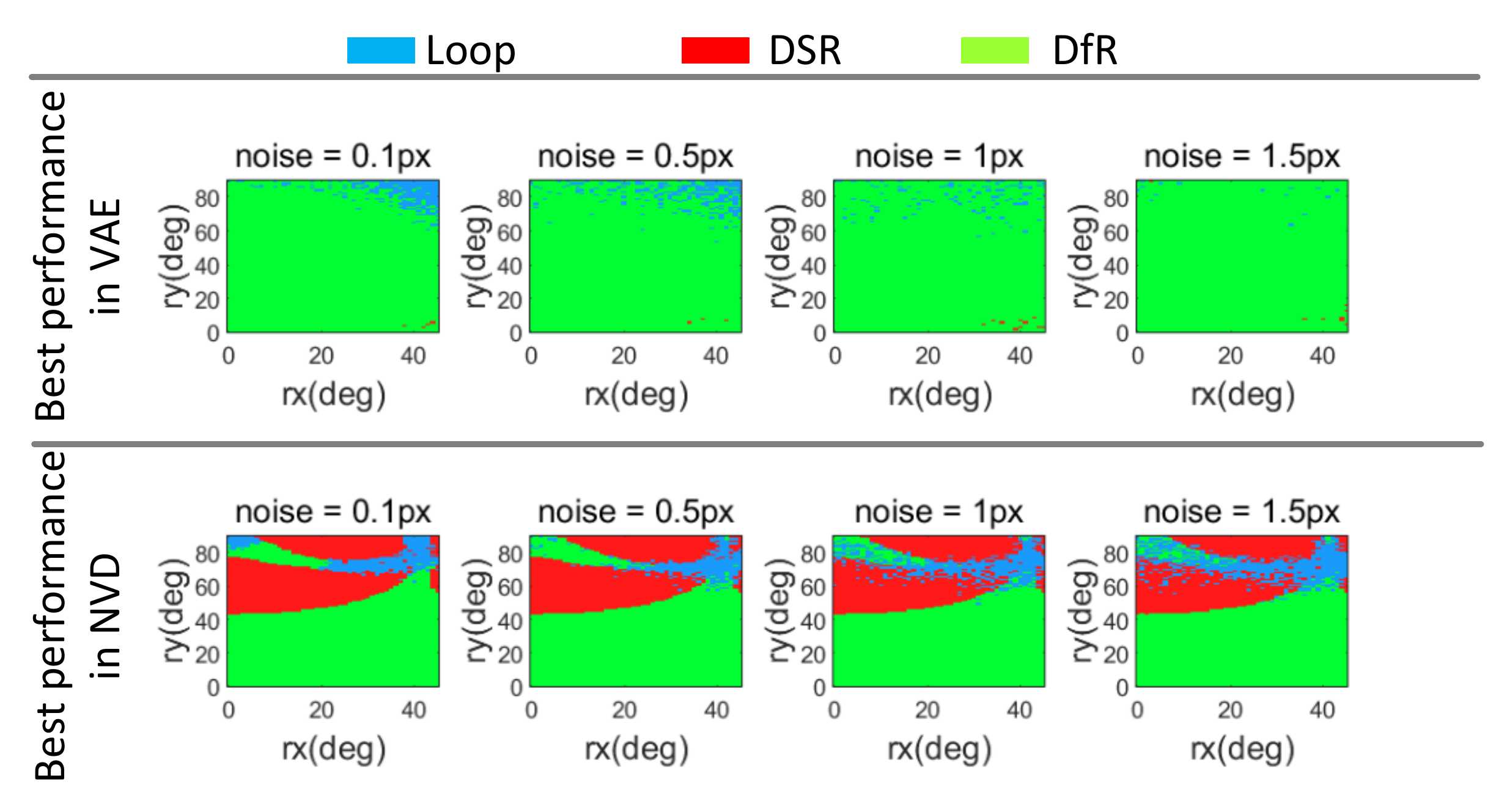}
    \caption{Best performance among \textit{Loop}, \textit{DSR} and \textit{DfR} in terms of VAE and NVD under different relative poses.}
    \label{fig:plot}
\end{figure}

\begin{figure*}[t]
    \centering
    \includegraphics[width=.9\linewidth]{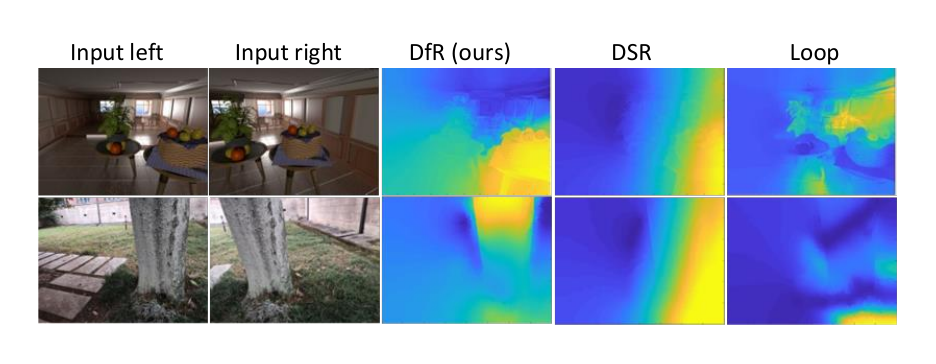}
    \caption{Comparisons of the depth maps recovered by \textit{Loop}, \textit{DSR} and \textit{DfR}.}
    \label{fig:real_data}
\end{figure*}

\section{Experiment}
\subsection{Synthetic Data}

\noindent \textbf{$\bullet$ Comparison Methods.} We compare the proposed method \textit{DfR} against to two state-of-the-art works:

    \textit{$\circ$ Loop~\cite{loop1999computing}:} Classical uncalibrated stereo rectification approach. 
    
    \textit{$\circ$ DSR~\cite{xiao2018dsr}:} Direct stereo rectification with small motion assumption.

    \textit{$\circ$ DfR:} The proposed method in this paper. 
    \quad \\
    
\noindent \textbf{$\bullet$ Metrics.} We evaluate the solutions by two metrics:

\textit{$\circ$ VAE:} Vertical alignment error defined in Eq.~(\ref{euqation:VAE}).  
 
\textit{$\circ$ NVD:} We measure the geometric distortion by a normalized vertex distance. Let $\mathbf{v_{1}}=[0\ 0\ 1],\mathbf{v_{2}}=[W-1\ 0\ 1],\mathbf{v_{3}}=[0\ H-1\ 1],\mathbf{v_{4}}=[W-1\ H-1\ 1]$ be the four vertices. Then 

  \begin{equation}\label{euq:NVD}
NVD=\frac{d_{1}+d_{2}+d_{3}+d_{4}}{\sqrt{W^{2}+H^{2}}}
\end{equation}

\noindent where $d_i$ is the distance from $\mathbf{v}_i$ to its rectified point.
 
\noindent \textbf{$\bullet$ Setting.} We generated two uncalibrated pin-hole cameras filming a cube scene under latitudinal motion. We set the radius as 1cm and varied the scene depth from 0.5m to 200m, the roll angle between two views from 0 to 45 degs, and the pitch angle from 0 to 90 degs. The results are obtained after averaging the errors over 50 trials at each setting. 

\noindent \textbf{$\bullet$ Results.} The results in Fig.~\ref{fig:box} show that the classical rectification method \textit{Loop}~\cite{loop1999computing} provides the most unstable estimation and huge error in VAE. While \textit{DSR} achieves much more robust results and higher accuracy. However, the proposed method \textit{DfR} outperforms \textit{Loop} and \textit{DSR} with a significant margin. Similarly, \textit{DfR} produces rectified results with much smaller and more stable NVD score than \textit{DSR} and \textit{Loop}, which indicates that the proposed method leads to less geometric distortion.  
Fig.~\ref{fig:plot} is to explore in more detail the effect of different conditions on the image rectification results. We mainly modify 3 conditions: the pitch angle ry of the rotating camera swinging left and right, the roll angle rx of the camera swinging up and down, and the image noise. For any points in Fig.~\ref{fig:plot}, the three conditions notes on the vertical axis, the horizontal axis, and the subheadings are shown. Under those conditions, the point is colored green if our proposed method is superior, blue if the method of Loop~\cite{loop1999computing} is superior, and red if the DSR~\cite{xiao2018dsr} is superior. \textit{DfR} achieves the best performance in both VAE and NVD under various stereo relative pose settings (roll and pitch).

\noindent \textbf{$\bullet$ Ablation Study.} We ablate the geometric distortion suppression strategy (Sec.~\ref{section:distortion_suppress}) used in the proposed \textit{DfR}. As shown in Fig.~\ref{fig:ablation}, Visually, \textit{DfR} produces significantly limited distortion over \textit{Loop} and \textit{DSR}. Quantitatively, \textit{Loop} provides the biggest NDV while \textit{DfR} achieves the smallest ones. It is vital to notice that the geometric distortion becomes larger when we remove the proposed suppression strategy.

\begin{table}[t]
\centering
\begin{tabular}{c | c} 
 \hline
  Methods & homographies estimation \\
 \hline
Loop~\cite{loop1999computing} & 14.638 ms\\  
 \hline
DSR~\cite{xiao2018dsr} & 0.135 ms\\ 
 \hline
DfR(Ours) & \textbf{0.038} ms\\ [1ex] 
 \hline
\end{tabular}
\caption{Running time of Loop\cite{loop1999computing}, DSR\cite{xiao2018dsr} and DfR.}
\label{table:time}
\end{table}

\noindent \textbf{$\bullet$ Running Time.} The experiment was run on a laptop with Intel I5 CPU. the running times are reported in Tab.~\ref{table:time} with 720×960 pix as input. The results show that \textit{DfR} achieves $4\times$ and $400\times$ speedup over \textit{DSR} and \textit{Loop}, respectively.

\subsection{Real Data}

To validate the effectiveness of realistic images, we use rotating surveillance cameras to collect 300 pairs of images in various scenarios. We use SIFT feature for detection and matching, followed by stereo rectification with \textit{Loop}, \textit{DSR}, and \textit{DfR}, respectively. Then we perform an SGM dense matching based on the three rectified results to extract the depth images. 
Results in Fig.~\ref{fig:real_data} show that \textit{Loop} and \textit{DSR} produce blurred depth maps while the proposed approach \textit{DfR} outputs clean and sharp ones instead. This verifies the effectiveness of \textit{DfR} in aligning the point matches into the same row.

\section{Conclusion}

This paper presents a novel image rectification solution to uncalibrated cameras with latitudinal motion assumption. 
The proposed DfR achieves high accuracy in alignment with minor geometric distortion. Extensive experiments demonstrate the effectiveness and efficiency of the proposed solution. The presented DfR can be applied as a pre-processing step to stereo matching for many applications, such as 3D visual surveillance and 3D perception of robotics.

\end{document}